\documentclass[runningheads]{llncs}
\usepackage{graphicx}

\usepackage{tikz}
\usepackage{comment}
\usepackage{amsmath,amssymb} 
\usepackage{color}
\usepackage{multirow}
\usepackage{floatrow}
\usepackage{hyperref}

\usepackage{lipsum}
\makeatletter
\def\@fnsymbol#1{\ensuremath{\ifcase#1\or *\or \dagger\or \ddagger\or
\mathsection\or \mathparagraph\or \|\or **\or \dagger\dagger
\or \ddagger\ddagger \else\@ctrerr\fi}}
\makeatother

\begin{document}

\renewcommand{\thefootnote}{\fnsymbol{footnote}}
\pagestyle{headings}
\mainmatter

\title{Actions as Moving Points}


\titlerunning{Actions as Moving Points}
%

\author{Yixuan Li\thanks{Yixuan Li and Zixu Wang contribute equally to this work.} \and
Zixu Wang$^*$ \and
Limin Wang\thanks{Corresponding author. This work is supported by Tencent AI Lab.} \and
Gangshan Wu
}
\authorrunning{Y. Li, Z. Wang et al.}
%
\institute{State Key Laboratory for Novel Software Technology, Nanjing University, China
\email{\{liyixxxuan,zixuwang1997\}@gmail.com, \{lmwang, gswu\}@nju.edu.cn}}

\maketitle
\begin{abstract}
  The existing action tubelet detectors often depend on heuristic anchor design and placement, which might be computationally expensive and sub-optimal for precise localization. In this paper, we present a conceptually simple, computationally efficient, and more precise action tubelet detection framework, termed as {\em MovingCenter Detector} (MOC-detector), by treating an action instance as a trajectory of moving points. Based on the insight that movement information could simplify and assist action tubelet detection, our MOC-detector is composed of three crucial head branches: (1) Center Branch for instance center detection and action recognition, (2) Movement Branch for movement estimation at adjacent frames to form trajectories of moving points, (3) Box Branch for spatial extent detection by directly regressing bounding box size at each estimated center. These three branches work together to generate the tubelet detection results, which could be further linked to yield video-level tubes with a matching strategy. Our MOC-detector outperforms the existing state-of-the-art methods for both metrics of frame-mAP and video-mAP on the JHMDB and UCF101-24 datasets. The performance gap is more evident for higher video IoU, demonstrating that our MOC-detector is particularly effective for more precise action detection. We provide the code at \href{https://github.com/MCG-NJU/MOC-Detector}{https://github.com/MCG-NJU/MOC-Detector}.
\end{abstract}

\section{Introduction}\label{introduction}

Spatio-temporal action detection is an important problem in video understanding, which aims to recognize all action instances present in a video and also localize them in both space and time. It has wide applications in many scenarios, such as video surveillance~\cite{videoSurveillance2,videoSurveillance1}, video captioning~\cite{videoCaptioning1,videoCaptioning2} and event detection~\cite{gan2015devnet}. Some early approaches~\cite{gkioxari2015finding,peng2016multi,SahaSSTC16,Wang0TG16,weinzaepfel2015learning,singh2017online} apply an action detector at each frame independently and then generate action tubes by linking these frame-wise detection results~\cite{gkioxari2015finding,peng2016multi,SahaSSTC16,Wang0TG16,singh2017online} or tracking one detection result~\cite{weinzaepfel2015learning} across time. These methods fail to well capture temporal information when conducting frame-level detection, and thus are less effective for detecting action tubes in reality. To address this issue, some approaches~\cite{saha2017amtnet,kalogeiton2017action,hou2017tube,yang2019step,zhao2019dance,song2019tacnet} try to perform action detection at the clip-level by exploiting short-term temporal information. In this sense, these methods input a sequence of frames and directly output detected tubelets (i.e., a short sequence of bounding boxes). This tubelet detection scheme yields a more principled and effective solution for video-based action detection and has shown promising results on standard benchmarks.

\begin{figure}[t]
  \centering
    \includegraphics[width=1\linewidth]{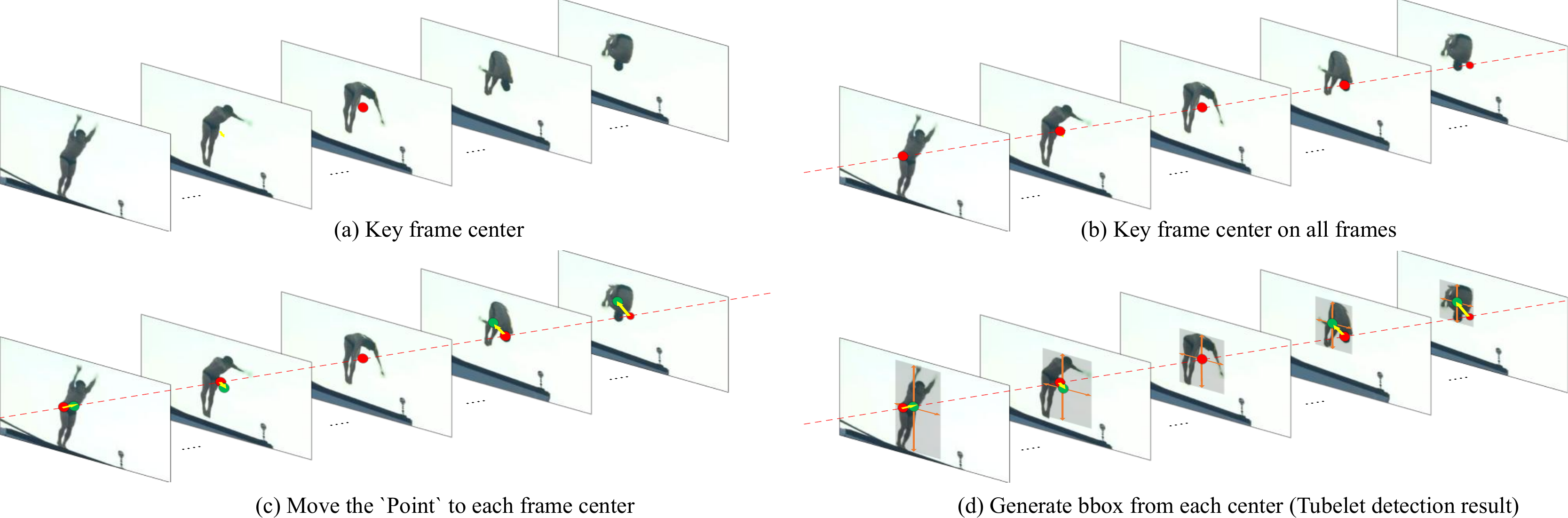}
    \caption{\textbf{Motivation Illustration.}
    We focus on devising an action tubelet detector from a short sequence. Movement information naturally describes human behavior, and each action instance could be viewed as a trajectory of {\em moving points}. In this view, action tubelet detector could be decomposed into three simple steps: (1) localizing the center point (red dots) at key frame (i.e., center frame), (2) estimating the movement at each frame with respect to the center point (yellow arrows), (3) regressing bounding box size at the calculated center point (green dots) for all frames.
    Best viewed in color and zoom in.
    }
    \label{fig:motivation}
  \end{figure}

The existing tubelet detection methods~\cite{saha2017amtnet,kalogeiton2017action,hou2017tube,yang2019step,zhao2019dance,song2019tacnet} are closely related with the current mainstream object detectors such as Faster R-CNN~\cite{ren2015faster} or SSD~\cite{liu2016ssd}, which operate on a huge number of pre-defined anchor boxes. Although these anchor-based object detectors have achieved success in image domains, they still suffer from critical issues such as being sensitive to hyper-parameters (e.g., box size, aspect ratio, and box number) and less efficient due to densely placed bounding boxes. These issues are more serious when adapting the anchor-based detection framework from images to videos. First, the number of possible tubelet anchors would grow dramatically when increasing clip duration, which imposes a great challenge for both training and inference. Second, it is generally required to devise more sophisticated anchor box placement and adjustment to consider the variation along the temporal dimension. In addition, these anchor-based methods directly extend 2D anchors along the temporal dimension which predefine each action instance as a cuboid across space and time. This assumption lacks the flexibility to well capture temporal coherence and correlation of adjacent frame-level bounding boxes.

Inspired by the recent advances in anchor-free object detection~\cite{YOLOV1,CornerNet,duan2019centernet,CenterNet,FCOS}, we present a {\bf conceptually simple, computationally efficient, and more precise} action tubelet detector in videos, termed as {\em MovingCenter detector} (MOC-detector). As shown in Figure~\ref{fig:motivation}, our detector presents a new tubelet detection scheme by treating each instance as a trajectory of {\em moving points}. In this sense, an action tubelet is represented by its center point in the key frame and offsets of other frames with respect to this center point. To determine the tubelet shape, we directly regress the bounding box size along the moving point trajectory on each frame. Our MOC-detector yields a fully convolutional one-stage tubelet detection scheme, which not only allows for more efficient training and inference but also could produce more precise detection results (as demonstrated in our experiments).

Specifically, our MOC detector decouples the task of tubelet detection into three sub-tasks: center detection, offset estimation and box regression. First, frames are fed into a 2D efficient backbone network for feature extraction. Then, we devise three separate branches: (1) Center Branch: detecting the action instance center and category; (2) Movement Branch: estimating the offsets of the current frame with respect to its center; (3) Box Branch: predicting bounding box size at the detected center point of each frame. This unique design enables three branches cooperate with each other to generate the tubelet detection results. Finally, we link these detected action tubelets across frames to yield long-range detection results following the common practice~\cite{kalogeiton2017action}. We perform experiments on two challenging action tube detection benchmarks of UCF101-24~\cite{UCF101} and JHMDB~\cite{JHMDB}. Our MOC-detector outperforms the existing state-of-the-art approaches for both frame-mAP and video-mAP on these two datasets, in particular for higher IoU criteria. Moreover, the fully convolutional nature of MOC detector yields a high detection efficiency of around 25FPS.

\section{Related Work}

\subsection{Object Detection}
\textbf{Anchor-based Object Detectors.}
Traditional one-stage~\cite{liu2016ssd,YOLOV1,lin2017focal} and two-stage object detectors~\cite{girshick2014rich,he2015spatial,girshick2015fast,ren2015faster} heavily relied on predefined anchor boxes. Two-stage object detectors like Faster-RCNN~\cite{ren2015faster}  and Cascade-RCNN~\cite{cai2018cascade} devised RPN to generate RoIs from a set of anchors in the first stage and handled classification and regression of each RoI in the second stage. By contrast, typical one-stage detectors utilized class-aware anchors and jointly predicted the categories and relative spatial offsets of objects, such as SSD~\cite{liu2016ssd}, YOLO~\cite{YOLOV1} and RetinaNet~\cite{lin2017focal}.

\textbf{Anchor-free Object Detectors.}
However, some recent works~\cite{FCOS,CenterNet,CornerNet,duan2019centernet,zhou2019bottom} have shown that the performance of anchor-free methods could be competitive with anchor-based detectors and such detectors also get rid of computation-intensive anchors and region-based CNN. CornerNet~\cite{CornerNet} detected object bounding box as a pair of corners, and grouped them to form the final detection. CenterNet~\cite{CenterNet} modeled an object as the center point of its bounding box and regressed its width and height to build the final
result.

\subsection{Spatio-temporal Action Detection}

\textbf{Frame-level Detector.}
Many efforts have been made to extend an image object detector to the task of action detection as frame-level action detectors~\cite{gkioxari2015finding,Wang0TG16,peng2016multi,SahaSSTC16,singh2017online,weinzaepfel2015learning}. After getting the frame detection, linking
algorithm is applied to generate final tubes~\cite{gkioxari2015finding,Wang0TG16,peng2016multi,SahaSSTC16,singh2017online} and Weinzaepfel et al.~\cite{weinzaepfel2015learning} utilized a tracking-by-detection method instead. Although flows are used to capture motion information, frame-level detection fails to fully utilize the
video's temporal information.

\textbf{Clip-level Detector.}
In order to model temporal information for detection, some clip-level approaches or action tubelet detectors~\cite{kalogeiton2017action,hou2017tube,yang2019step,li2018recurrent,zhao2019dance,song2019tacnet} have been proposed. ACT~\cite{kalogeiton2017action} took a short sequence of frames and output tubelets which were regressed from anchor cuboids.
STEP~\cite{yang2019step} proposed a progressive method to refine the proposals over a few steps to solve the large displacement problem and utilized longer temporal information. Some methods~\cite{hou2017tube,li2018recurrent} linked frame or tubelet proposals first to generate tubes proposal and then did classification.

These approaches are all based on anchor-based object detectors, whose design might be sensitive to anchor design and computationally cost due to large numbers of anchor boxes. Instead, we try to design an anchor-free action tubelet detector by treating each action instance as a trajectory of moving points. Experimental results demonstrate that our proposed action tubelet detector is effective for spatio-temporal action detection, in particular for the high video IoU.

\section{Approach}

\begin{figure}[t]
  \centering
    \includegraphics[width=1\linewidth]{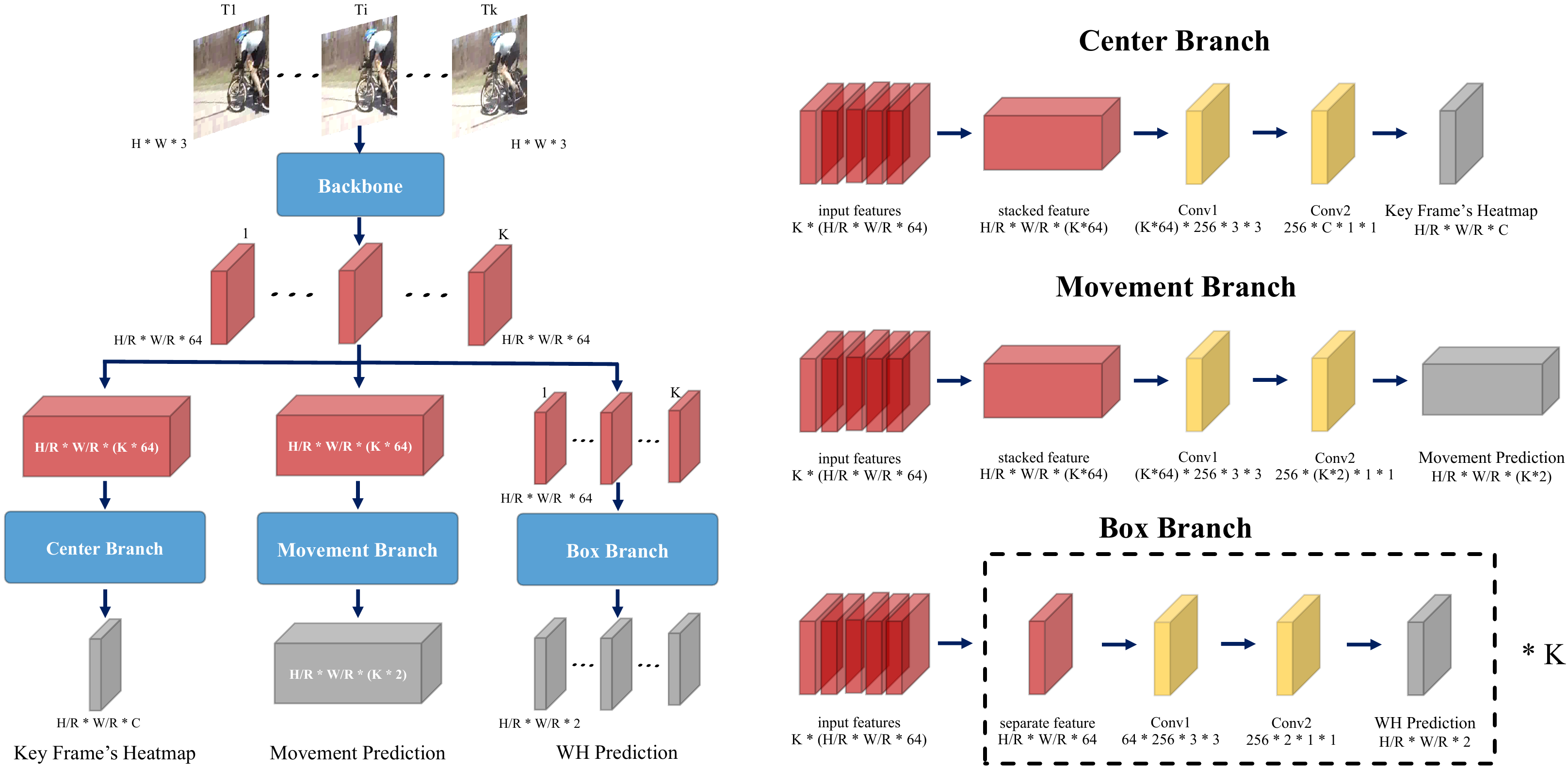}
    \caption{{\bf Pipeline of MOC-detector}.
    In the left, we present the overall MOC-detector framework. The red cuboids represent the extracted features, the blue boxes denote the backbone or detection head, and the gray cuboids are detection results produced by the Center Branch, the Movement Branch, the Box Branch.
    In the right, we show the detailed design of each branch. Each branch consists of a sequence of one 3*3 conv layer, one ReLu layer and one 1*1 conv layer, which is presented as yellow cuboids. The parameters of convolution are input channel, output channel, convolution kernel height, convolution kernel width.
    }
    \label{fig:pipeline}
  \end{figure}

{\bf Overview.} Action tubelet detection aims at localizing a short sequence of bounding boxes from an input clip and recognizing its action category as well. We present a new tubelet detector, coined as {MovingCenter detector} (MOC-detector), by viewing an action instance as a trajectory of moving points. As shown in Figure~\ref{fig:pipeline}, in our MOC-detector, we take a set of consecutive frames as input and separately feed them into an efficient 2D backbone to extract frame-level features. Then, we design three head branches to perform tubelet detection in an anchor-free manner. The first branch is Center Branch, which is defined on the center (key) frame. This Center Branch localizes the tubelet center and recognizes its action category. The second branch is Movement Branch, which is defined over all frames. This Movement Branch tries to relate adjacent frames to predict the center movement along the temporal dimension. The estimated movement would propagate the center point from key frame to other frames to generate a trajectory. The third branch is Box Branch, which operates on the detected center points of all frames. This branch focuses on determining the spatial extent of the detected action instance at each frame, by directly regressing the height and width of the bounding box. These three branches collaborate together to yield tubelet detection from a short clip, which will be further linked to form action tube detection in a long untrimmed video by following a common linking strategy~\cite{kalogeiton2017action}. We will first give a short description of the backbone design, and then provide technical details of three branches and the linking algorithm in the following subsections.

{\bf Backbone.} In our MOC-detector, we input $K$ frames and each frame is with the resolution of $W \times H$.  First $K$ frames are fed into a 2D backbone network sequentially to generate a feature volume $\mathbf{f} \in \mathbb{R}^{K \times\frac{W}{R} \times \frac{H}{R} \times B}$. $R$ is the spatial downsample ratio and $B$ denotes channel number. To keep the full temporal information for subsequent detection, we do not perform any downsampling over the temporal dimension. Specifically, we choose DLA-34~\cite{DLA} architecture as our MOC-detector feature backbone following CenterNet~\cite{CenterNet}.
This architecture employs an encoder-decoder architecture to extract features for each frame. The spatial downsampling ratio $R$ is 4 and the channel number $B$ is 64. The extracted features are shared by three head branches.
Next we will present the technical details of these head branches.

\subsection{Center Branch: Detect Center at Key Frame}\label{CenterBranch}
The Center Branch aims at detecting the action instance center in the key frame (i.e., center frame) and recognizing its category based on the extracted video features. Temporal information is important for action recognition, and thereby we design a temporal module to estimate the action center and recognize its class by concatenating multi-frame feature maps along channel dimension. Specifically, based on the video feature representation $\mathbf{f} \in \mathbb{R}^{\frac{W}{R} \times \frac{H}{R} \times (K \times B)}$, we estimate a center heatmap $\hat{L} \in [0,1]^{\frac{W}{R}\times\frac{H}{R}\times C}$ for the key frame. The $C$ is the number of action classes. The value of $\hat{L}_{x,y,c}$ represents the likelihood of detecting an action instance of class $c$ at location $(x, y)$, and higher value indicates a stronger possibility. Specifically, we employ a standard convolution operation to estimate the center heatmap in a fully convolutional manner.

\textbf{Training.} We train the Center Branch following the common dense prediction setting~\cite{CornerNet,CenterNet}. For $i^{th}$ action instance, we represent its center as key frame's bounding box center and utilize center's position for each action category as the ground truth label $(x_{c_i},y_{c_i})$. We generate the ground truth heatmap $L\in[0,1]^{\frac{W}{R}\times\frac{H}{R}\times C}$ using a
 Gaussian kernel which produces the soft heatmap groundtruth $L_{x,y,c_i}=\exp(-\frac{(x-x_{c_i})^2+(y-y_{c_i})^2}{2\sigma_p^2})$. For other class (i.e., $c\neq c_i$), we set the heatmap $L_{x,y,c}=0$.
The $\sigma_p$ is adaptive to instance size and we choose the maximum when two Gaussian of the same category overlap. We choose the training objective, which is a variant of focal loss~\cite{lin2017focal}, as follows:

\begin{equation}
  \begin{split}
  \ell_{\mathrm{center}}=-\frac{1}{n}
  \sum_{x,y,c}\left\{
      \begin{array}{lc}
      (1-\hat{L}_{xyc})^\alpha \log(\hat{L}_{xyc}) & \mathrm{if} \ L_{xyc}=1 \\
      (1-L_{xyc})^{\beta}(\hat{L}_{xyc})^\alpha \log(1-\hat{L}_{xyc}) & \mathrm{otherwise}
     \end{array}
  \right.
  \end{split}
  \end{equation}
where $n$ is the number of ground truth instances and $\alpha$ and $\beta$ are hyper-parameters of the focal loss~\cite{lin2017focal}. We set $\alpha=2$ and $\beta=4$ following~\cite{CornerNet,CenterNet} in our experiments. It indicates that this focal loss is able to deal with the imbalanced training issue effectively~\cite{lin2017focal}.

\textbf{Inference.} After the training, the Center Branch could be deployed in tubelet detection for localizing action instance center and recognizing its category. Specifically, we detect all local peaks which are equal to or greater than their 8-connected neighbors in the estimated heatmap $\hat{L}$ for each class independently. And then keep the top $N$ peaks from all categories as candidate centers with tubelet scores.
Following~\cite{CenterNet}, we set $N$ as 100 and detailed ablation studies will be provided in the appendix A.

\subsection{Movement Branch: Move Center Temporally}\label{sec:movementBranch}
\label{sec:movement}

The Movement Branch tries to relate adjacent frames to predict the movement of the action instance center along the temporal dimension. Similar to Center Branch, Movement Branch also employs temporal information to regress the center offsets of current frame with respect to key frame. Specifically, Movement Branch takes stacked feature representation as input and outputs a movement prediction map $\hat{M} \in \mathbb{R}^{\frac{W}{R}\times\frac{H}{R}\times(K\times 2)}$. $2K$ channels represent center movements from key frame to current frames in $X$ and $Y$ directions. Given the key frame center $(\hat{x}_{key},\hat{y}_{key})$, $\hat{M}_{\hat{x}_{key},\hat{y}_{key},2j:2j+2}$ encodes center movement at $j^{th}$ frame.

\textbf{Training.} The ground truth tubelet of $i^{th}$ action instance is \begin{footnotesize}
$[(x_{tl}^1,y_{tl}^1,x_{br}^1,y_{br}^1),$

\noindent $...,(x_{tl}^j,y_{tl}^j,x_{br}^j,y_{br}^j),...,(x_{tl}^K,y_{tl}^K,x_{br}^K,y_{br}^K)]$
\end{footnotesize}
, where subscript $tl$ and $br$ represent top-left and bottom-right points of bounding boxes, respectively. Let $k$ be the key frame index, and the $i^{th}$ action instance center at key frame is defined as follows:
\begin{equation}
  (x^{key}_{i},y^{key}_{i})=(\lfloor(x_{tl}^{k}+x_{br}^{k})/2\rfloor,\lfloor(y_{tl}^{k}+y_{br}^{k})/2\rfloor).
\end{equation}
We could compute the bounding box center $(x_{i}^j,y_{i}^j)$ of $i^{th}$ instance at $j^{th}$ frame as follows:
\begin{equation}
  (x_{i}^{j},y_{i}^{j})=((x_{tl}^{j}+x_{br}^{j})/2,(y_{tl}^{j}+y_{br}^{j})/2).
  \end{equation}
Then, the ground truth movement of the $i^{th}$ action instance is calculated as follows:
\begin{equation}
m_i=(x_{i}^{1}-x^{key}_{i} ,y_{i}^{1}-y_{i}^{key},...,x_{i}^{K}-x_{i}^{key},y_{i}^{K}-y_{i}^{key}).
\end{equation}
For the training of Movement Branch, we optimize the movement map $\hat{M}$ only at the key frame center location and use the $\ell_1$ loss as follows:
\begin{equation}
  \ell_{\mathrm{movement}}=\frac{1}{n}\sum_{i=1}^{n}|\hat{M}_{x^{key}_i,y^{key}_i}-m_i|.
\end{equation}

\textbf{Inference.} After the Movement Branch training and given $N$ detected action centers $\{(\hat{x}_i,\hat{y}_i)| i \in \{1, 2, \cdots, N \}\}$ from Center Branch, we obtain a set of movement vector $\{\hat{M}_{\hat{x}_i,\hat{y}_i}|i\in \{1, 2, \cdots, N \}\}$ for all detected action instance. Based on the results of Movement Branch and Center Branch, we could easily generate a trajectory set $T=\{T_i| i\in \{1, 2, \cdots, N \} \}$, and for the detected action center $(\hat{x}_i,\hat{y}_i)$, its trajectory of moving points is calculated as follows:
\begin{equation}
    T_i=(\hat{x}_i,\hat{y}_i) + [\hat{M}_{\hat{x}_i,\hat{y}_i,0:2} , \hat{M}_{\hat{x}_i,\hat{y}_i,2:4}, \cdots, \hat{M}_{\hat{x}_i,\hat{y}_i,2K-2:2K}].
\end{equation}

\subsection{Box Branch: Determine Spatial Extent}

The Box Branch is the last step of tubelet detection and focuses on determining the spatial extent of the action instance. Unlike Center Branch and Movement Branch, we assume box detection only depends on the current frame and temporal information will not benefit the class-agnostic bounding box generation. We will provide the ablation study in the appendix B. In this sense, this branch could be performed in a frame-wise manner. Specifically, Box Branch inputs the single frame's feature $\mathbf{f}^{j} \in \mathbb{R}^{\frac{W}{R}\times\frac{H}{R} \times B}$ and generates a size prediction map $\hat{S}^j\in \mathbb{R}^{\frac{W}{R}\times\frac{H}{R} \times 2}$  for the $j^{th}$ frame to directly estimate the bounding box size (i.e., width and height). Note that the Box Branch is shared across K frames.

\textbf{Training.}
The ground truth bbox size of $i^{th}$ action instance at $j^{th}$ frame can be represented as follows:
\begin{equation}
    s_i^j = (x^{j}_{br} - x^j_{tl}, y^j_{br} - y^j_{tl}).
\end{equation}
With this ground truth bounding box size, we optimize the Box Branch at the center points of all frames for each tubelet with $\ell_1$ Loss as follows:
\begin{equation}
  \ell_{\mathrm{box}}=\frac{1}{n}\sum_{i=1}^{n} \sum_{j=1}^{K} |\hat{S}_{p_i^j}^j-s_i^j|.
\end{equation}
Note that the $p_{i}^{j}$ is the $i^{th}$ instance ground truth center at $j^{th}$ frame.
So the overall training objective of our MOC-detector is
\begin{equation}\label{loss}
\ell=\ell_{\mathrm{center}}+a\ell_{\mathrm{movement}}+b\ell_{\mathrm{box}},
\end{equation}
where we set a=1 and b=0.1 in all our experiments. Detailed ablation studies will be provided in the appendix A.

\textbf{Inference.} Now, we are ready to generate the tubelet detection results. based on center trajectories $T$ from Movement Branch and
 size prediction heatmap $\hat{S}$ for each location produced by this branch. For $j^{th}$ point in trajectory $T_i$, we use $(T_{x},T_{y})$ to denote its coordinates, and (w,h) to denote Box Branch size output $\hat{S}$ at specific location. Then, the bounding box for this point is calculated as:
 \begin{small}
 \begin{equation}
  (T_{x}-w/2,T_{y}-h/2, T_{x}+w/2,T_{y}+h/2).
 \end{equation}
\end{small}

\subsection{Tubelet Linking}\label{link}
After getting the clip-level detection results, we link these tubelets into final tubes across time. As our main goal is to propose a new tubelet detector, we use the same linking algorithm as~\cite{kalogeiton2017action} for fair comparison. Given a video, MOC extracts tubelets and keeps the top 10 as candidates for each sequence of K frames with stride 1 across time, which are linked into the final tubes in a tubelet by tubelet manner. \textbf{Initialization:} In the first frame, every candidate starts a new link. At a given frame, candidates which are not assigned to any existing links start new links. \textbf{Linking:} At a given frame, we extend the existing links with one of the tubelet candidates starting at this frame in descending order of links' scores. The score of a link is the average score of tubelets in this link. One candidate can only be assigned to one existing link when it meets three conditions: (1) the candidate is not selected by other links, (2) the overlap between link and candidate is greater than a threshold $\tau$, (3) the candidate $t$ has the highest score. \textbf{Termination:} An existing link stops if it has not been extended in consecutive K frames. We build an action tube for each link, whose score is the average score of tubelets in the link. For each frame in the link, we average the bbox coordinates of tubelets containing that frame. Initialization and termination determine tubes' temporal extents. Tubes with low confidence and short duration are abandoned. As this linking algorithm is online, MOC can be applied for online video stream.

\section{Experiments}

\subsection{Experimental Setup}\label{Implementation details}
{\bf Datasets and Metrics.} We perform experiments on the UCF101-24~\cite{UCF101} and JHMDB~\cite{JHMDB} datasets. UCF101-24~\cite{UCF101} consists of 3207 temporally untrimmed videos from 24 sports classes. Following the common setting~\cite{peng2016multi,kalogeiton2017action}, we report the action detection performance for the first split only. JHMDB~\cite{JHMDB} consists of 928 temporally trimmed videos from 21 action classes. We report results averaged over three splits following the common setting~\cite{peng2016multi,kalogeiton2017action}. AVA~\cite{gu2018ava} is a larger dataset for action detection but only contains a single-frame action instance annotation for each 3s clip, which concentrates on detecting actions on a single key frame. Thus, AVA is not suitable to verify the effectiveness of tubelet action detectors. Following~\cite{weinzaepfel2015learning,gkioxari2015finding,kalogeiton2017action}, we utilize frame mAP and video mAP to evaluate detection accuracy.

\noindent{\bf Implementation Details.}
We choose the DLA34~\cite{DLA} as our backbone with COCO~\cite{lin2014microsoft} pretrain and ImageNet~\cite{deng2009imagenet} pretrain. We provide MOC results with COCO pretrain without extra explanation. For a fair comparison, we provide two-stream results on two datasets with both COCO pretrain and ImageNet pretrain in Section~\ref{SOTA}. The frame is resized to $288 \times 288$. The spatial downsample ratio $R$ is set to 4 and the resulted feature map size is $72 \times 72$. During training, we use the same data augmentation as~\cite{kalogeiton2017action} to the whole video: photometric transformation, scale jittering, and location jittering. We use Adam with a learning rate 5e-4 to optimize the overall objective. The learning rate adjusts to convergence on the validation set and it decreases by a factor of 10 when performance saturates. The iteration maximum is set to 12 epochs on UCF101-24~\cite{UCF101} and 20 epochs on JHMDB~\cite{JHMDB}.

\subsection{Ablation Studies}\label{Ablation Study}
For efficient exploration, we perform experiments only using RGB input modality, COCO pretrain, and K as 5 without extra explanation. Without special specified, we use exactly the same training strategy in this subsection.

\subsubsection{Effectiveness of Movement Branch.}

\begin{figure}[t]
    \includegraphics[width=1\textwidth]{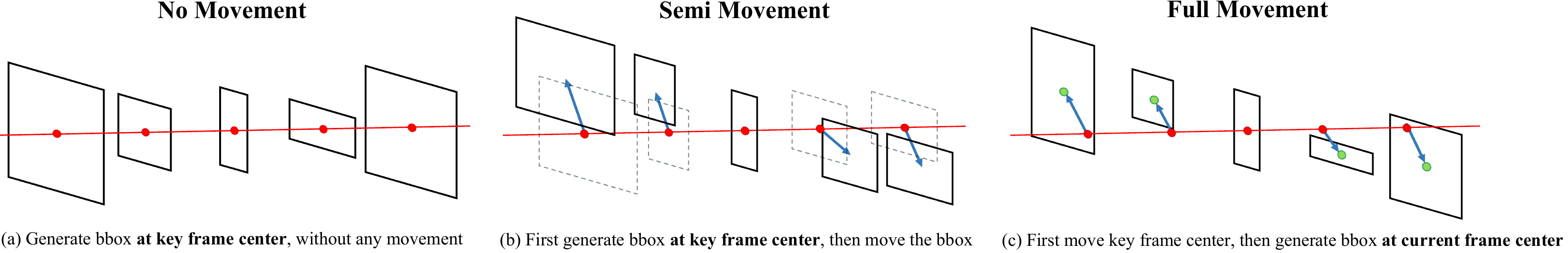}
    \caption{\textbf{Illustration of Three Movement Strategies.}
    Note that the arrow represents moving according to Movement Branch prediction, the red dot represents the key frame center and the
    green dot represents the current frame center, which is localized by moving key frame center according to Movement Branch prediction.
    }
    \label{fig:core}
\end{figure}

In MOC, Movement Branch impacts on both bbox's location and size. Movement Branch moves key frame center to other frames to locate bbox center, named as Move Center strategy. Box Branch estimates bbox size on the current frame center, which is located by Movement Branch not the same with key frame, named as Bbox Align strategy. To explore the effectiveness of Movement Branch, we compare MOC with other two detector designs, called as {\em No Movement} and {\em Semi Movement}. We set the tubelet length $K = 5$ in all detection designs with the same training strategy. As shown in Figure~\ref{fig:core}, \textbf{No Movement} directly removes the Movement Branch and just generates the bounding box for each frame at the same location with key frame center. \textbf{Semi Movement} first generates the bounding box for each frame at the same location with key frame center, and then moves the generated box in each frame according to Movement Branch prediction.  \textbf{Full Movement (MOC)} first moves the key frame center to the current frame center according to Movement Branch prediction, and then Box Branch generates the bounding box for each frame at its own center. The difference between Full Movement and Semi Movement is that they generate the bounding box at different locations: one at the real center, and the other at the fixed key frame center. The results are summarized in Table~\ref{tbl:nomove}.

\begin{table}[t]
  \begin{center}
  \caption{Exploration study on MOC detector design with various combinations of movement strategies on UCF101-24.}
  \label{tbl:nomove}
  \resizebox{\linewidth}{!}{
  \begin{tabular}{c|cc|c|cccc}
  \hline
  \multirow{2}{*}{Method}&\multicolumn{2}{c|}{Strategy}&\multirow{2}{*}{F-mAP@0.5 (\%)}&\multicolumn{4}{c}{Video-mAP (\%)} \\
  \cline{2-3}\cline{5-8}
  \iftrue
  &Move Center &Bbox Align& & @0.2 & @0.5 & @0.75 & 0.5:0.95\\
  \hline
  No Movement&&&68.22&68.91&37.77&19.94&19.27\\
  Semi Movement&\checkmark&&69.78&76.63&48.82&\bf27.05&26.09\\
  Full Movement (MOC)&\checkmark&\checkmark&\bf 71.63&\bf 77.74&\bf 49.55&27.04&\bf 26.09\\
  \fi
  \hline
  \end{tabular}
  }
  \end{center}
\end{table}

 {\em First}, we observe that the performance gap between No Movement and Semi Movement is 1.56\% for frame mAP@0.5 and 11.05\% for video mAP@0.5. We find that the Movement Branch has a relatively small influence on frame mAP, but contributes much to improve the video mAP. Frame mAP measures the detection quality in a single frame without tubelet linking while video mAP measures the tube-level detection quality involving tubelet linking. Small movement in short tubelet doesn't harm frame mAP dramatically but accumulating these subtle errors in the linking process will seriously harm the video-level detection. So it demonstrates that the movement information is important for improving video mAP. {\em Second}, we can see that Full Movement performs slightly better than Semi Movement for both video mAP and frame mAP. Without Bbox Align, Box Branch estimates bbox size at key frame center for all frames, which causes a small performance drop with MOC. This small gap implies that Box Branch is relatively robust to the box center and estimating bbox size at small shifted location only brings a very slight performance difference.

\subsubsection{Study on Movement Branch Design.}\label{Study on Movement Branch design}
\begin{table}[t]
  \caption{Exploration study on the Movement Branch design on UCF101-24~\cite{UCF101}. Note that our MOC-detector adopts the Center Movement.}
    \label{tbl:move}
  \begin{center}
  \resizebox{0.8\linewidth}{!}{
  \begin{tabular}{c|c|cccc}
  \hline
  \multirow{2}{*}{Method}&\multirow{2}{*}{F-mAP@0.5 (\%)}&\multicolumn{4}{c}{Video-mAP (\%)} \\
  \cline{3-6}
  \iftrue
  & & @0.2 & @0.5 & @0.75 & 0.5:0.95\\
  \hline
  Flow Guided Movement&69.38&75.17&42.28&22.26&21.16\\
  Cost Volume Movement&69.63&72.56&43.67&21.68&22.46\\
  Accumulated Movement&69.40&75.03&46.19&24.67&23.80\\
  Center Movement&\bf 71.63&\bf 77.74&\bf 49.55&\bf 27.04&\bf26.09\\
  \fi
  \hline
  \end{tabular}
  }
  \end{center}
\end{table}

In practice, in order to find an efficient way to capture center movements, we implement Movement Branch in several different ways.
The first one is {\em Flow Guided Movement} strategy which utilizes optical flow between adjacent frames to move action instance center.  The second strategy, {\em Cost Volume Movement}, is to directly compute the movement offset by constructing cost volume between key frame and current frame following~\cite{zhao2018recognize}, but this explicit computing fails to yield better results and is slower due to the constructing of cost volume. The third one is {\em Accumulated Movement} strategy which predicts center movement between consecutive frames instead of with respect to key frame. The fourth strategy, {\em Center Movement}, is to employ 3D convolutional operation to directly regress the offsets of the current frame with respect to key frame as illustrated in Section~\ref{sec:movement}. The results are reported in Table~\ref{tbl:move}.

We notice that the simple Center Movement performs best and choose it as Movement Branch design in our MOC-detector, which directly employs a 3D convolution to regress key frame center movement for all frames as a whole. We will analyze the fail reason for other three designs. For {\em Flow Guided Movement}, (i) Flow is not accurate and just represents pixel movement, while {\em Center Movement} is supervised by box movement. (ii) Accumulating adjacent flow to generate trajectory will enlarge error. For the {\em Cost Volume Movement}, (i) We explicitly calculate the correlation of the current frame with respect to key frame. When regressing the movement of the current frame, it only depends on the current correlation map. However, when directly regressing movement with 3D convolutions, the movement information of each frame will depend on all frames, which might contribute to more accurate estimation. (ii) As cost volume calculation and offset aggregation involve a correlation without extra parameters, it is observed that the convergence is much harder than {\em Center Movement}. For {\em Accumulated Movement}, this strategy also causes the issue of error accumulation and is more sensitive to the training and inference consistency. In this sense, the ground truth movement is calculated at the real bounding box center during training, while for inference, the current frame center is estimated from Movement Branch and might not be so precise, so that {\em Accumulated Movement} would bring large displacement to the ground truth.

\subsubsection{Study on Input Sequence Duration.}\label{Input Sequence Duration}

\begin{table}[t]
  \begin{center}
  \caption{Exploration study on the tubelet duration $K$ on UCF101-24.}
  \label{tbl:k}
  \resizebox{0.75\linewidth}{!}{
  \begin{tabular}{c|c|cccc}
  \hline
  \multirow{2}{*}{Tubelet Duration}&\multirow{2}{*}{F-mAP@0.5 (\%)}&\multicolumn{4}{c}{Video-mAP (\%)}
  \\
  \cline{3-6}
  \iftrue
  & & @0.2 & @0.5 & @0.75 & 0.5:0.95\\
  \hline
  $K = 1$&68.33&65.47&31.50&15.12&15.54\\
  $K = 3$&69.94&75.83&45.94&24.94&23.84\\
  $K = 5$&71.63&77.74&49.55&27.04&26.09\\
  $K = 7$&\bf 73.14&\bf78.81&\bf 51.02&\bf 27.05&\bf 26.51\\
  $K = 9$& 72.17& 77.94&50.16&26.26& 26.07\\
  \fi
  \hline
  \end{tabular}
  }
  \end{center}
\end{table}

The temporal length $K$ of the input clip is an important parameter in our MOC-detector. In this study, we report the RGB stream performance of MOC on UCF101-24~\cite{UCF101} by varying $K$ from 1 to 9 and the experiment results are summarized in Table~\ref{tbl:k}. We reduce the training batch size for K = 7 and K = 9 due to GPU memory limitation.

{\em First}, we notice that when $K = 1$, our  MOC-detector reduces to the frame-level detector which obtains the worst performance, in particular for video mAP. This confirms the common assumption that frame-level action detector lacks consideration of temporal information for action recognition and thus it is worse than those tubelet detectors, which agrees with our basic motivation of designing an action tubelet detector. {\em Second}, we see that the detection performance will increase as we vary $K$ from 1 to 7 and the performance gap becomes smaller when comparing $K = 5$ and $K = 7$. From $K = 7$ to $K = 9$, detection performance drops because predicting movement is harder for longer input length.  According to the results, we set $K = 7$ in our MOC.

\subsection{Comparison with the State of the Art}\label{SOTA}

\begin{table}[t]
  \begin{center}
    \caption{Comparison with the state of the art on JHMDB (trimmed) and UCF101-24 (untrimmed). Ours (MOC)${}^{\dagger}$ is pretrained on ImageNet~\cite{deng2009imagenet} and \textbf{Ours (MOC)} is pretrained on COCO~\cite{lin2014microsoft}.}
    \label{table:state}
  \resizebox{1\textwidth}{!}{
  \begin{tabular}{c|c|cccc|c|cccc}
  \hline
  \multirow{3}{*}{Method} &\multicolumn{5}{c|}{JHMDB} & \multicolumn{5}{c}{UCF101-24 }  \\
  \cline{2-11}
  &  \multirow{2}{*}{Frame-mAP@0.5 (\%)}& \multicolumn{4}{c|}{Video-mAP (\%)}  & \multirow{2}{*}{Frame-mAP@0.5 (\%)}   & \multicolumn{4}{c}{Video-mAP (\%)} \\
  \cline{3-6}
  \cline{8-11}
  \iftrue
  & & @0.2 & @0.5 & @0.75 & 0.5:0.95 & & @0.2 & @0.5 & @0.75 & 0.5:0.95\\
  \hline
  \hline
  \multicolumn{11}{l}{\textbf{2D Backbone}}\\
  \hline
  Saha {\em et al.} 2016 ~\cite{SahaSSTC16}& - & 72.6 & 71.5 & 43.3 & 40.0& - & 66.7 & 35.9 & 7.9 & 14.4\\
  Peng {\em et al.} 2016~\cite{peng2016multi}& 58.5 & 74.3 & 73.1 & - & - &  39.9 & 42.3 & - & - & - \\
  Singh {\em et al.} 2017~\cite{singh2017online} &  - & 73.8 & 72.0 & 44.5 & 41.6& - & 73.5 & 46.3 & 15.0 & 20.4\\
  Kalogeiton {\em et al.} 2017~\cite{kalogeiton2017action} & 65.7 & 74.2 & 73.7 & 52.1 & 44.8 &  69.5 & 76.5 & 49.2 & 19.7 & 23.4  \\
  Yang {\em et al.} 2019~\cite{yang2019step}& - & - & - & - & - &  75.0 & 76.6 & - & - & - \\
  Song {\em et al.} 2019~\cite{song2019tacnet} &  65.5 & 74.1 & 73.4 & 52.5 & 44.8 & 72.1 & 77.5 & 52.9 & 21.8 & 24.1 \\
  Zhao {\em et al.} 2019~\cite{zhao2019dance} & - & - & 74.7 & 53.3 & 45.0 & - & 78.5 & 50.3 & 22.2 & 24.5 \\
  Ours (MOC)${}^{\dagger}$ & 68.0 & 76.2 & 75.4 & 68.5 & 54.0 & 76.9 & 81.3 & \bf{54.4} & 29.5 & \bf{28.4} \\
  {\bf Ours (MOC)}&  {\bf 70.8} & {\bf 77.3} & {\bf 77.2} & {\bf 71.7} & {\bf 59.1} & \bf{78.0} & \bf{82.8} & 53.8 & \bf{29.6} & 28.3 \\
  \hline
  \hline
  \multicolumn{11}{l}{\textbf{3D Backbone}}\\
  \hline
  Hou {\em et al.} 2017~\cite{hou2017tube} (C3D)& 61.3 & \bf 78.4 & 76.9 & - & -& 41.4 & 47.1 & - & - & -  \\
  Gu {\em et al.} 2018~\cite{gu2018ava} (I3D)& 73.3 & - & 78.6  & - & - &  \bf 76.3 & - & \bf 59.9 &  - & - \\
  Sun {\em et al.} 2018~\cite{SunSVMSS18} (S3D-G)& \bf 77.9 &   - &\bf 80.1 &  - & - & - & - & - & - & - \\
  \fi
  \hline
  \end{tabular}
  }
  \end{center}
\end{table}

 Finally, we compare our MOC with the existing state-of-the-art methods on the trimmed JHMDB dataset and the untrimmed UCF101-24 dataset in Table~\ref{table:state}. For a fair comparison, we also report two-stream results with ImageNet pretrain.

 Our MOC gains similar performance on UCF101-24 for ImageNet pretrain and COCO pretrain, while COCO pretrain obviously improves MOC's performance on JHMDB because JHMDB is quite small and sensitive to the pretrain model.
 Our method significantly outperforms those frame-level action detectors~\cite{SahaSSTC16,peng2016multi,singh2017online} both for frame-mAP and video-mAP, which perform action detection at each frame independently without capturing temporal information. \cite{kalogeiton2017action,yang2019step,zhao2019dance,song2019tacnet} are all tubelet detectors, our MOC outperforms them for all metrics on both datasets, and the improvement is more evident for high IoU video mAP. This result confirms that our anchor-free MOC detector is more effective for localizing precise tubelets from clips than those anchor-based detectors, which might be ascribed to the flexibility and continuity of MOC detector by directly regressing tubelet shape.
 Our methods get comparable performance to those 3D backbone based methods~\cite{hou2017tube,gu2018ava,SunSVMSS18}. These methods usually divide action detection into two steps: person detection (ResNet50-based Faster RCNN~\cite{ren2015faster} pretrained on ImageNet), and action classification (I3D~\cite{carreira2017quo}/S3D-G~\cite{xie2018rethinking} pretrained on Kinetics~\cite{carreira2017quo} + ROI pooling), and fail to provide a simple unified action detection framework.

\begin{figure}
  \centering
    \includegraphics[width=1\linewidth]{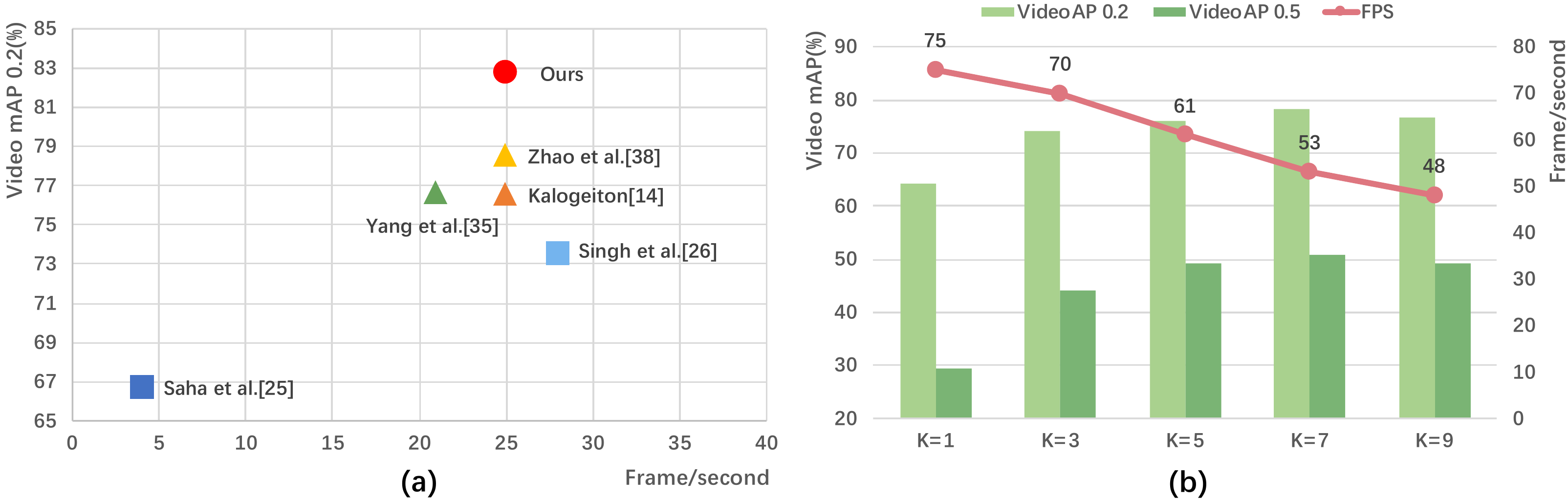}
    \caption{{\bf Runtime Comparison and Analysis}. (a) Comparison with other methods. Two-stream results following ACT~\cite{kalogeiton2017action}'s setting. (b) The detection accuracy (green bars) and speeds (red dots) of MOC's online setting.}
  \label{fig:runtime}
\end{figure}
\begin{figure}
  \centering
    \includegraphics[width=0.8\linewidth]{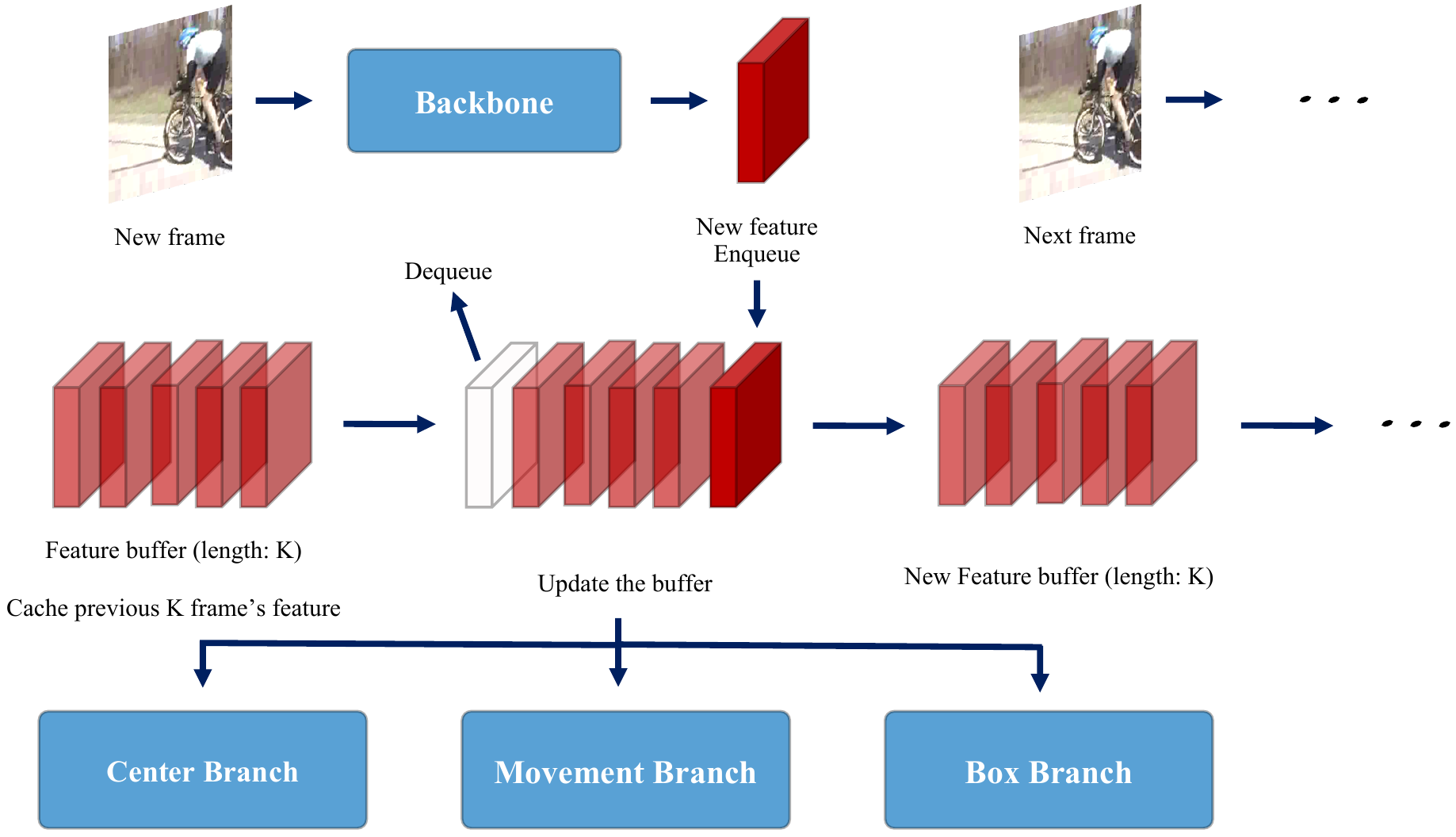}
    \caption{{\bf Process of Handling Online Video Stream}.}
  \label{fig:online}
\end{figure}
\subsection{Runtime Analysis}
Following ACT~\cite{kalogeiton2017action}, we evaluate MOC's two-stream offline speed on a single GPU without including flow extraction time and MOC reaches 25 fps. In Figure~\ref{fig:runtime}(a), we compare MOC with some existing methods which have reported their speed in the original paper. ~\cite{yang2019step,zhao2019dance,kalogeiton2017action} are all action tubelet detectors and our MOC gains more accurate detection results with comparable speed.
Our MOC can be applied for processing online real-time video stream, which is shown in Figure~\ref{fig:online}. To simulate online video stream, we set batch size as 1. Since the backbone feature can be extracted only once, we save previous K-1 frames' features in a buffer. When getting a new frame, MOC's backbone first extracts its feature and combines with the previous K-1 frames' features in the buffer. Then MOC's three branches generate tubelet detections based on
these features. After that, update the buffer by adding current frame's feature for subsequent detection.
For online testing, we only input RGB as optical flow extraction is quite expensive and the results are reported in Figure~\ref{fig:runtime}(b).  We see that our MOC is quite efficient in online testing and it reaches 53 FPS for K = 7.

\subsection{Visualization}

\begin{figure}[t]
  \centering
    \includegraphics[width=1\linewidth]{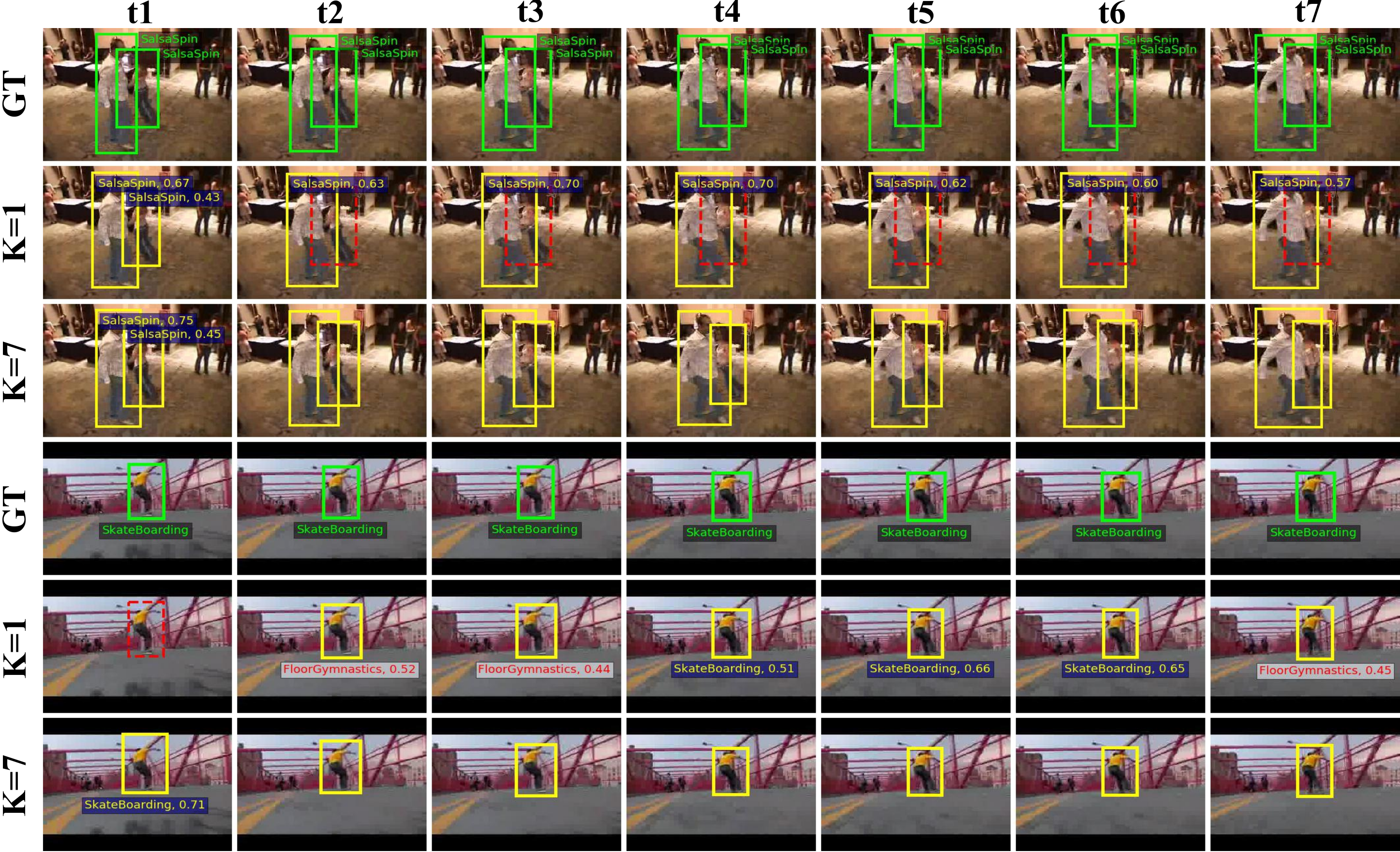}
    \caption{{\bf Examples of Per-frame (K = 1) and Tubelet (K = 7) Detection}. The yellow color boxes present detection results, whose categories and scores are provided beside. Yellow categories represent correct and red ones represent wrong. Red dashed boxes represent missed actors. Green boxes and categories are the ground truth. MOC generates one score and category for one tubelet and we mark these in the first frame of the tubelet. Note that we set the visualization threshold as 0.4.}
  \label{fig:K_vision}
\end{figure}

In Figure~\ref{fig:K_vision}, we give some qualitative examples to compare the performance between tubelet duration K = 1 and K = 7. Comparison between the second row and the third row shows that our tubelet detector leads to less missed detection results and localizes action more accurately owing to offset constraint in the same tubelet. What's more, comparison between the fifth and the sixth row presents that our tubelet detector can reduce classification error because some actions can not be discriminated by just looking one frame.

\section{Conclusion and Future Work}
In this paper, we have presented an action tubelet detector, termed as MOC, by treating each action instance as a trajectory of moving points and directly regressing bounding box size at estimated center points of all frames. As demonstrated on two challenging datasets, the MOC-detector has brought a new state-of-the-art with both metrics of frame mAP and video mAP, while maintaining a reasonable computational cost. The superior performance is largely ascribed to the unique design of three branches and their cooperative modeling ability to perform tubelet detection. In the future, based on the proposed MOC-detector, we try to extend its framework to longer-term modeling and model action boundary in the temporal dimension, thus contributing to spatio-temporal action detection in longer continuous video streams.

\noindent {\bf Acknowledgements}. This work is supported by Tencent AI Lab Rhino-Bird Focused Research Program (No. JR202025), the National Science Foundation of China (No. 61921006), Program for Innovative Talents and Entrepreneur in Jiangsu Province, and Collaborative Innovation Center of Novel Software Technology and Industrialization.

\section*{Appendix A: Study on Hyper-parameters}\label{A}
\begin{figure}
    \centering
      \includegraphics[width=0.8\linewidth]{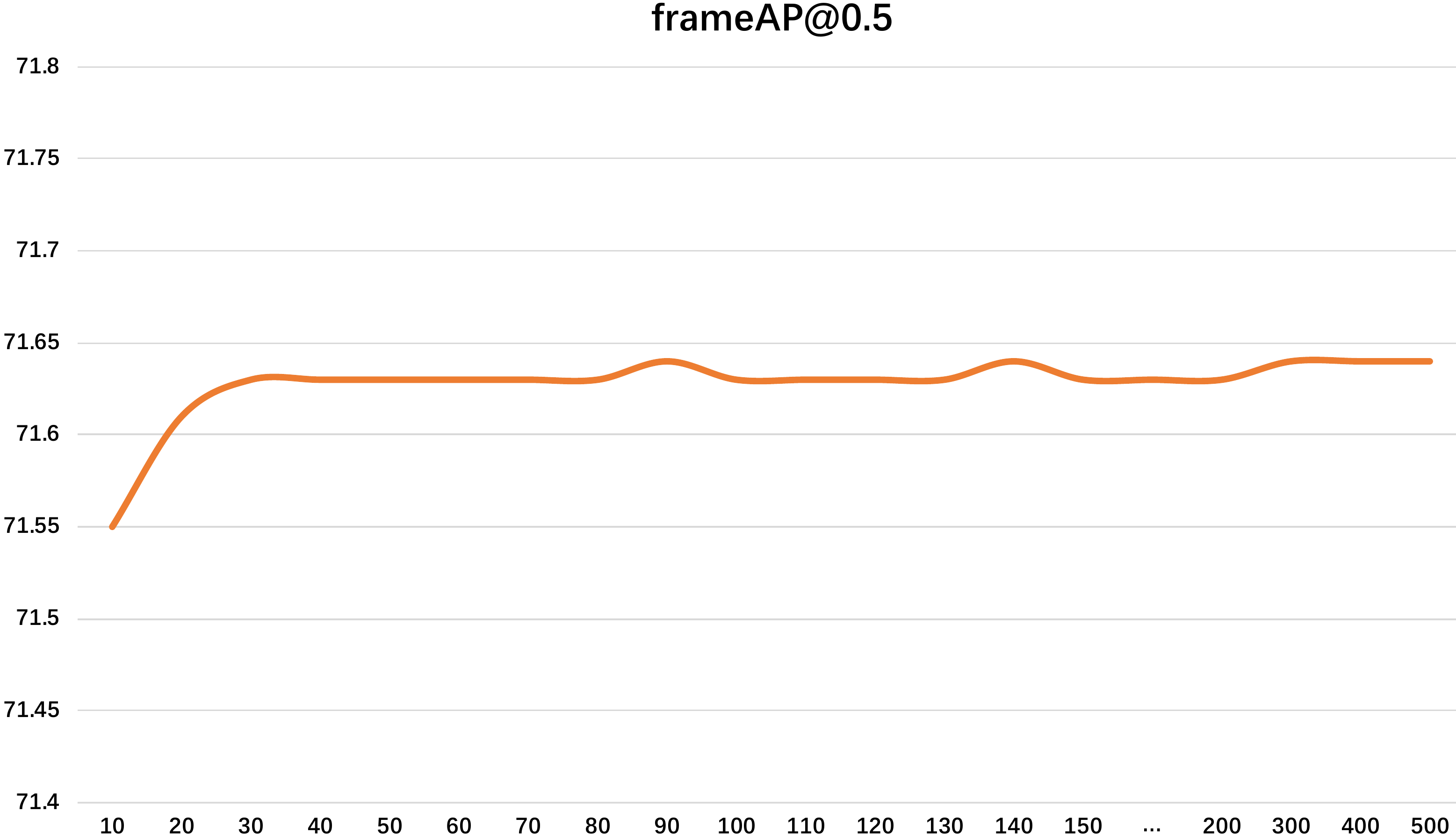}
      \caption{{\bf Study on N}. FrameAP@0.5 result on UCF101-24~\cite{UCF101} with tubelet length K=5 and only RGB input.}
    \label{fig:N}
\end{figure}

\noindent\textbf{N in Center Branch.}
During inference, Center Branch keeps top N instances from all categories after max pooling operation, which is indicated in paper's Section 3.1. We follow CenterNet~\cite{CenterNet}, which is an anchor-free object detector and set N as 100.
As shown in Figure~\ref{fig:N}, we can see that the detection result is robust to N and changes slightly after 40.

\begin{figure}
    \centering
      \includegraphics[width=0.6\linewidth]{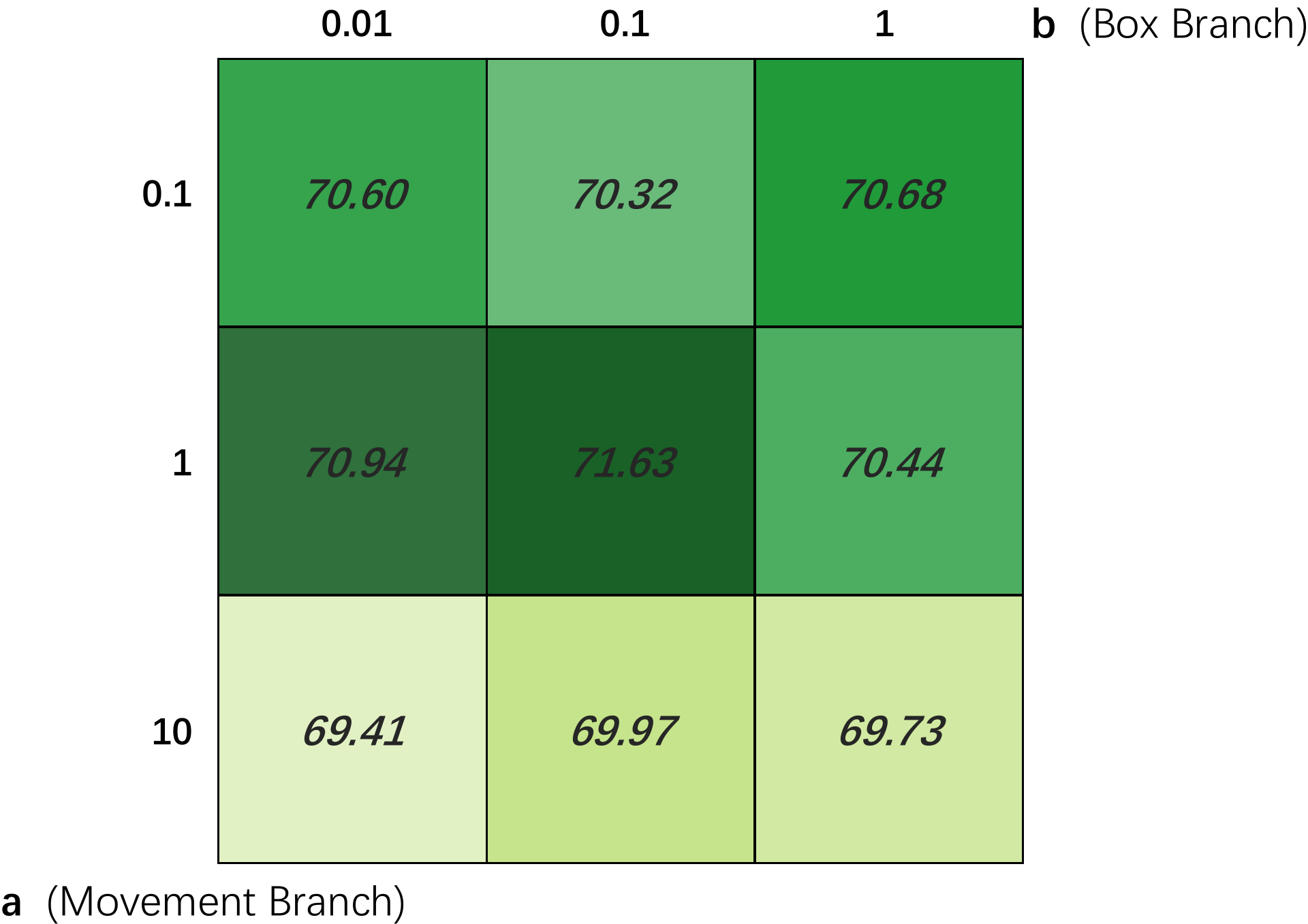}
      \caption{{\bf Study on a and b}. FrameAP@0.5 result on UCF101-24~\cite{UCF101} with tubelet length K=5 and only RGB input.
      }
    \label{fig:ab}
\end{figure}
\noindent\textbf{a and b in Loss Function.}
Paper's Equation (9) is MOC’s training objective consisting of three branches’ loss. As shown in Figure~\ref{fig:ab}, we have a linear search on a and b with tubelet length K=5 and only RGB input. We can see that a=1, b=0.1 performs best.

\section*{Appendix B: More exploration on Box Branch}
\begin{table}
  \caption{Exploration study on the Box Branch design with only RGB as input and $K = 5$. Note that union means stacking feature together to add temporal information into the bbox estimation and separate (MOC) estimates bbox separately for each frame.}
    \label{tbl:box}
  \begin{center}
  \begin{tabular}{c|c|cccc}
  \hline
  \multirow{2}{*}{Method}&\multirow{2}{*}{F-mAP@0.5 (\%)}&\multicolumn{4}{c}{Video-mAP (\%)} \\
  \cline{3-6}
  \iftrue
  & & @0.2 & @0.5 & @0.75 & 0.5:0.95\\
  \hline
  union&70.41&76.54&49.14&26.61&\bf26.14\\
  separate(MOC)&\bf 71.63&\bf 77.74&\bf 49.55&\bf 27.04&26.09\\
  \fi
  \hline
  \end{tabular}
  \end{center}
\end{table}

Previously, we tried to add temporal information into the bbox estimation by stacking features across time as input, which is as same as Movement Branch. As shown in Table~\ref{tbl:box}, the performance drops after adding temporal information. It indicates that a single frame is sufficient for the bbox detection.

\section*{Appendix C: Error Analysis}
\begin{figure}
    \centering
      \includegraphics[width=0.8\linewidth]{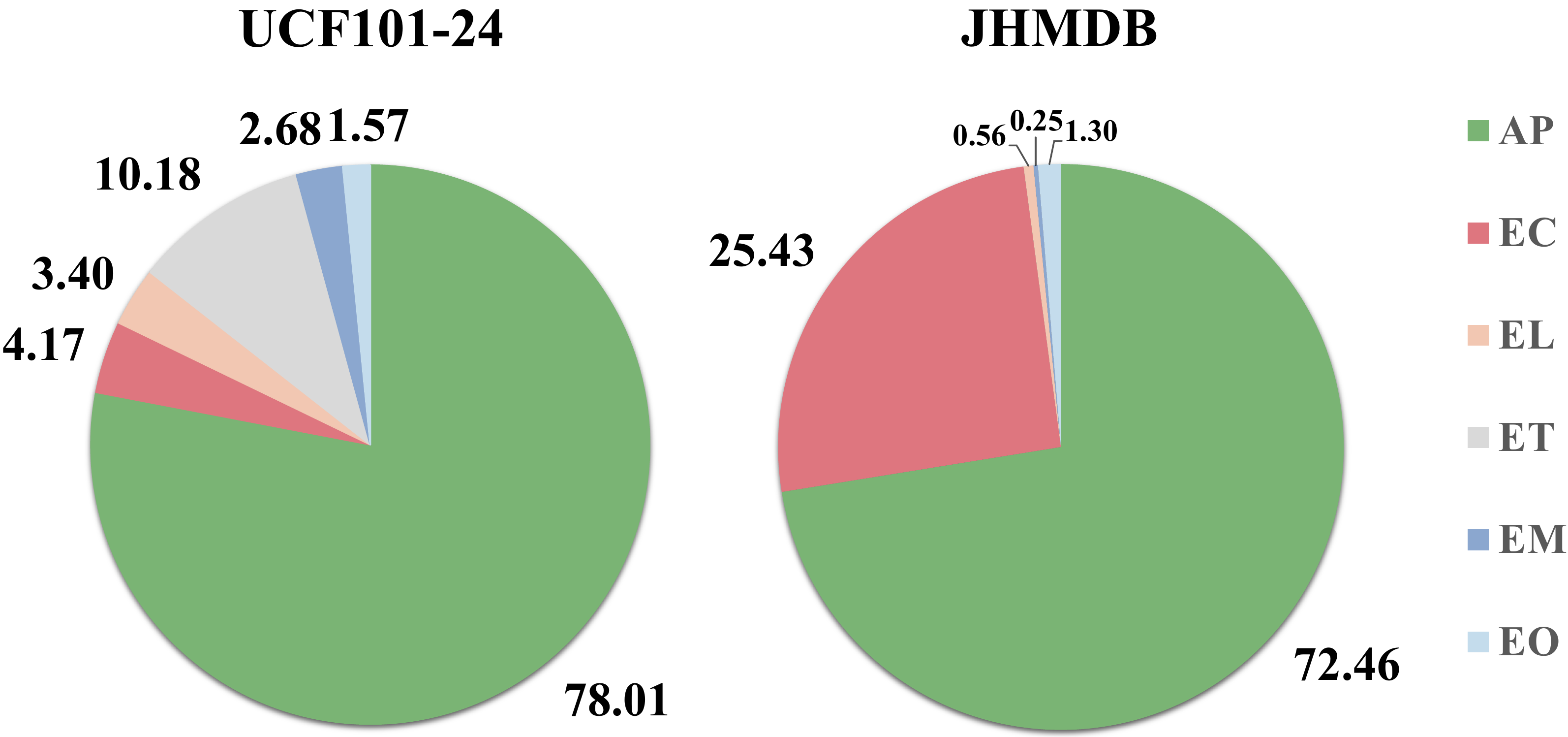}
      \caption{{\bf Error analysis on UCF101-24~\cite{UCF101} and JHMDB~\cite{JHMDB} (only split 1)}. We report the detection error results according to
      five categories: (1) classification error $E_C$, (2) localization error $E_L$, (3) time error $E_T$, (4) missed detection $E_M$, and (5) other error $E_O$. The green part represents the correct detection. With tubelet length K = 7 and two-stream fusion.
       }
    \label{fig:error}
\end{figure}
\begin{figure}
\centering
    \includegraphics[width=0.8\textwidth]{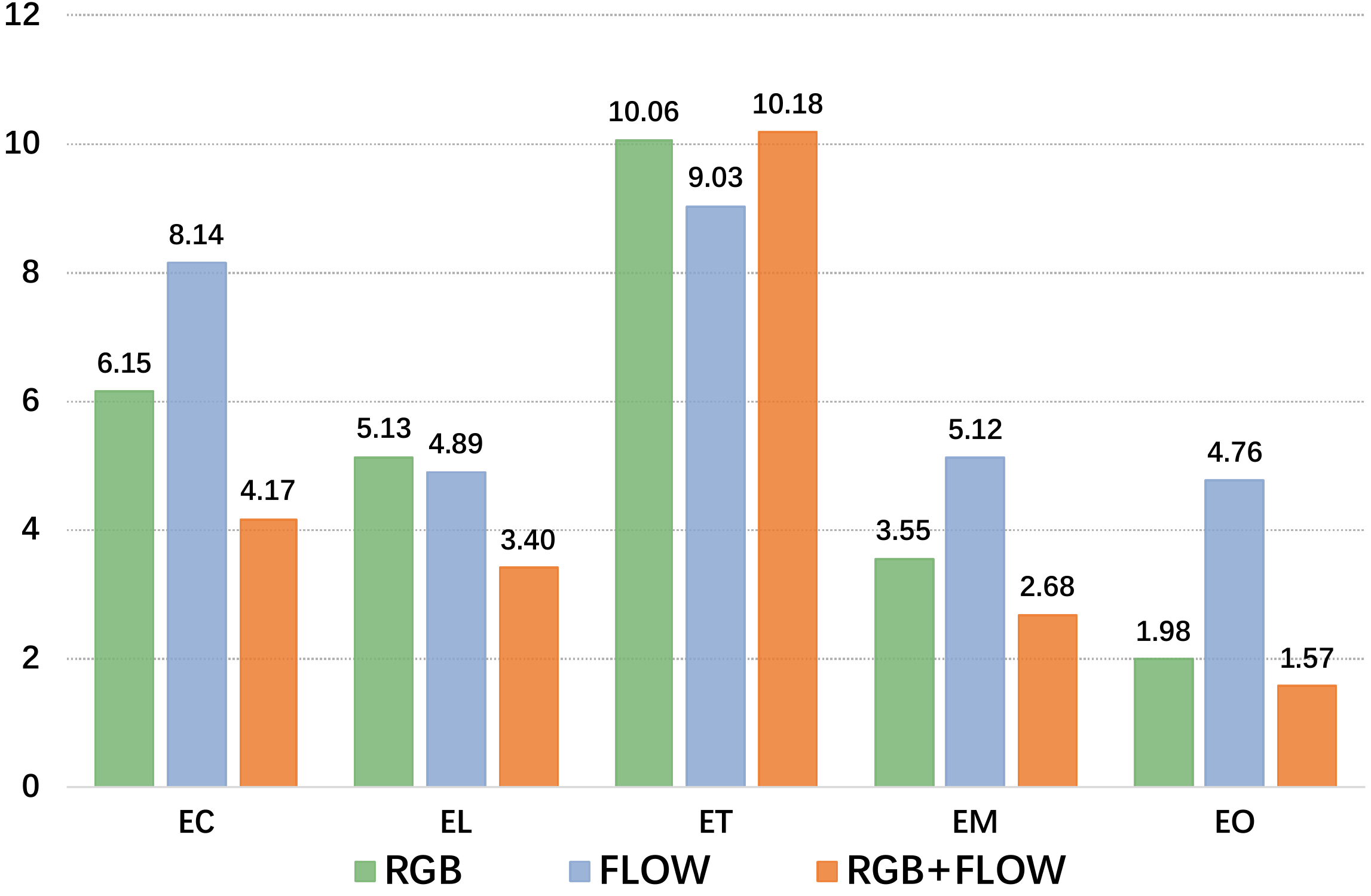}
    \caption{{\bf Error Analysis with Two-stream Fusion}. We report the detection error results according to five categories by changing input: (1) classification error $E_C$, (2) localization error $E_L$, (3) time error $E_T$, (4) missed detection $E_M$, and (5) other error $E_O$. With tubelet length K = 7 and two-stream fusion on UCF101-24~\cite{UCF101}.}
    \label{fig:error-fusion}
\end{figure}
In this section, following~\cite{kalogeiton2017action}, we conduct an error analysis on the frame mAP to better explore our proposed MOC-detector. In particular, we investigate five kinds of tubelet detection error: (1) classification error $E_C$: the detection IoU is greater than 0.5 with the ground-truth box of another action class. (2) localization error $E_L$: the detection class is correct in a frame but the bounding box IoU with ground truth is less than $0.5$. (3) time error $E_T$: the detection in the untrimmed video covers the frame that doesn't belong to the temporal extent of the current action instance. (4) missed detection error $E_M$: cannot detect out a ground truth box. (5) other error $E_O$: the detection appears in a frame without the class and has IoU less than 0.5 with the ground truth bounding box of other classes.

We present error analysis on the untrimmed dataset UCF101-24~\cite{UCF101} and the trimmed dataset JHMDB~\cite{JHMDB} (only split 1) with tubelet length $K=7$ and two-stream fusion. As shown in Figure~\ref{fig:error}, we find the major error is $E_T$, time error (10.18\%), for the untrimmed dataset UCF101-24~\cite{UCF101} and $E_C$, classification error (25.43\%), for the trimmed dataset JHMDB~\cite{JHMDB}. Although our MOC-detector has achieved state-of-art on both datasets, we will try to extend this framework to model longer temporal information to improve classification accuracy and model action boundary in the temporal dimension to eliminate time error.

We also visualize error analysis with two-stream fusion on UCF101-24~\cite{UCF101} and the results are reported in Figure~\ref{fig:error-fusion}. Note that we set tubelet length $K$ as 7. First, spatial stream performs obviously better than the temporal stream for classification error and missed detection, owing to its richer information. Second, two-stream fusion improves the performance except for time error, which shows that two-stream fusion harms temporal localization.

\section*{Appendix D: More Results on JHMDB}
\begin{table}
  \begin{center}
    \caption{Comparison with Gu et al.~\cite{gu2018ava} and Sun et al.~\cite{SunSVMSS18}  on JHMDB~\cite{JHMDB} (3 splits) with tubelet length K=7 and two stream fusion. Ours (MOC) ${}^{\dagger}$ is pretrained on ImageNet~\cite{deng2009imagenet} , Ours (MOC)${}^{\dagger}{}^{\dagger}$ is pretrained on COCO~\cite{lin2014microsoft} and Ours (MOC)${}^{\dagger}{}^{\dagger}{}^{\dagger}$ is pretrained on UCF101-24~\cite{UCF101} for action detection.}
    \label{table}
  \resizebox{1\textwidth}{!}{
  \begin{tabular}{c|c|c|cccc}
  \hline
  \multirow{3}{*}{Method}&\multirow{3}{*}{GFLOPs} &\multicolumn{5}{c}{JHMDB} \\
  \cline{3-7}&
  &  \multirow{2}{*}{Frame-mAP@0.5 (\%)}& \multicolumn{4}{c}{Video-mAP (\%)}  \\
  \cline{4-7}
  \iftrue
 & & & @0.2 & @0.5 & @0.75 & 0.5:0.95 \\
  \hline
  Ours (MOC)${}^{\dagger}$ &29.4& 68.0 & 76.2 & 75.4 & 68.5 & 54.0 \\
 Ours (MOC)${}^{\dagger}{}^{\dagger}$& 29.4& 70.8 & 77.3 &  77.2 & 71.7 & 59.1 \\
Ours (MOC)${}^{\dagger}{}^{\dagger}{}^{\dagger}$ &29.4& 74.0 & \bf 80.7 & \bf 80.5 & \bf 75.0 & \bf 60.2 \\
  \hline
  \hline
  Gu {\em et al.} 2018~\cite{gu2018ava} (I3D)& \textgreater91.0&73.3 & - & 78.6  & - & -  \\
  Sun {\em et al.} 2018~\cite{SunSVMSS18} (S3D-G)&\textgreater65.5& \bf 77.9 &   - &80.1 &  - & -  \\
  \fi
  \hline
  \end{tabular}
  }
  \end{center}
\end{table}

Our MOC is a one stage tubelet detector with 2D backbone. We compare it with two-stage detectors with 3D backbone~\cite{gu2018ava,SunSVMSS18} in paper’s Section 4.3, which perform comparably with us on UCF101-24~\cite{UCF101} while better than ours on JHMDB~\cite{JHMDB}.

JHMDB~\cite{JHMDB} is really small and sensitive to the pre-train model. For fair comparison with 2D backbone methods in paper's Section 4.3, we just provide results with ImageNet~\cite{deng2009imagenet} pretrain and COCO~\cite{lin2014microsoft} pretrain. But Gu et al~\cite{gu2018ava} and Sun et al.~\cite{SunSVMSS18} both pretrain 3D backbone on Kinetics~\cite{carreira2017quo}, which is a large-scale video classification dataset and always boosts task results especially on small datasets. We pretrain our MOC on UCF101-24~\cite{UCF101} for action detection in Table~\ref{table}, which outperforms Gu et al.~\cite{gu2018ava} for all metrics with saving more than 3 times computation cost and performs comparably with Sun et al.~\cite{SunSVMSS18} with saving more than 2 times computation cost. Note that Gu et al.~\cite{gu2018ava} and Sun et al.~\cite{SunSVMSS18} do not provide implementation code, so we just roughly estimate the backbone computation for each frame's detection result, whose input is 20 frames with resolution of 320*400. For Gu et al.~\cite{gu2018ava}, we calculate ResNet50 (conv4)~\cite{he2016deep} for action localization and I3D (Mixed\_4e)~\cite{carreira2017quo} for classification. For Sun et al.~\cite{SunSVMSS18} (Base Model), we calculate ResNet50 (conv4)~\cite{he2016deep} for action localization and S3D-G~\cite{xie2018rethinking} for classification. For our MOC, we calculate the whole computation cost for each frame detection result. For fair comparison, we only use RGB as input to estimate GFLOPs for all methods.

%
%
\bibliographystyle{splncs04}
\bibliography{egbib}
\end{document}